\documentclass{article}

\usepackage{PRIMEarxiv}

\usepackage[utf8]{inputenc} 
\usepackage[T1]{fontenc}    
\usepackage{hyperref}       
\usepackage{url}            
\usepackage{booktabs}       
\usepackage{amsfonts}       
\usepackage{nicefrac}       
\usepackage{microtype}      
\usepackage{fancyhdr}       
\usepackage{graphicx}       
\graphicspath{{media/}}     
\usepackage{cite}
\usepackage{amsmath,amssymb,amsfonts}
\usepackage{algorithmic}
\usepackage{algorithm}
\usepackage{graphicx}
\usepackage{caption}
\usepackage{subcaption}
\usepackage{textcomp}
\usepackage{xcolor}
\usepackage{epstopdf}
\usepackage{balance}
\usepackage{soul}

\definecolor{qquestion}{RGB}{3, 175, 30}
\definecolor{ffix}{RGB}{195, 6, 2}
\definecolor{llongterm}{RGB}{74, 80, 133}

\def\BibTeX{{\rm B\kern-.05em{\sc i\kern-.025em b}\kern-.08em
    T\kern-.1667em\lower.7ex\hbox{E}\kern-.125emX}}
\title{Iterative Forgetting: Online Data Stream Regression using Database-inspired Adaptive Granulation
\thanks{This material is based upon work supported by
the National Science Foundation under Grant No.
1932138. Any opinions, findings, and conclusions or
recommendations expressed in this material are those
of the authors and do not necessarily reflect the views
of the National Science Foundation.} 
}

\author{
  Niket Kathiriya \\
  Computer Science Department \\
  University of Massachusetts Lowell \\
  \texttt{Niket\_Kathiriya@student.uml.edu} \\
   \And
  Hossein Haeri \\
  Mechanical Engineering Department \\
  University of Massachusetts Lowell \\
  \texttt{Hossein\_Haeri@uml.edu} \\
   \And 
  Cindy Chen \\
  Computer Science Department \\
  University of Massachusetts Lowell \\
  \texttt{Cindy\_Chen@uml.edu} \\
   \And   
  Kshitij Jerath \\
  Mechanical Engineering Department \\
  University of Massachusetts Lowell \\
  \texttt{Kshitij\_Jerath@uml.edu} \\  
}

\begin{document}
\maketitle

\begin{abstract}

Many modern systems, such as financial, transportation and telecommunications systems, are time-sensitive in the sense that they demand low-latency predictions for real-time decision-making.
Such systems often have to contend with continuous unbounded data streams as well as concept drift, which are challenging requirements that traditional regression techniques are unable to cater to. There exists a need to create novel data stream regression methods that can handle these scenarios.
We present a database-inspired datastream regression model that (a) uses inspiration from R*-trees to create granules from incoming datastreams such that relevant information is retained, (b) iteratively forgets granules whose information is deemed to be outdated, thus maintaining a list of only recent, relevant granules, and (c) uses the recent data and granules to provide low-latency predictions. The R*-tree-inspired approach also makes the algorithm amenable to integration with database systems. Our experiments demonstrate that the ability of this method to discard data produces a significant order-of-magnitude improvement in latency \textit{and} training time when evaluated against the most accurate state-of-the-art algorithms, while the R*-tree-inspired granulation technique provides competitively accurate predictions. 

\end{abstract}


\section{Introduction}
Our fast-paced and highly interconnected world is comprised of highly dynamic, uncertain, and time-sensitive systems that operate and play a crucial role across domains such as finance, healthcare, transportation, and telecommunications, to name a few. Often, such systems also generate a continuous flow of the data referred to as data streams \cite{AGRAHARI2022}. To maintain a desired, optimal level of performance, many of these systems need to operate in ultra-low latency conditions with response times on the order of a few microseconds to a few milliseconds \cite{Nasrallah2019}.
Traditional regression techniques often struggle to keep up with the velocity and volume of data streams, necessitating approaches that can handle the dynamic nature of the data while ensuring timely predictions. Within the realm of data stream regression techniques, managing the memory requirements associated with the incoming data streams while producing accurate predictions is in itself a major challenge.

We propose a novel approach for data stream regression that offers several advantages over the state-of-the-art such as: (a) faster data stream processing and predictions, (b) systematic identification and removal of expired data, leading to reduced memory requirements, and (c) facilitating batch processing for data streams, which can be useful in applications with large number of data stream sources that result in high frequency of incoming data. Perhaps equally as importantly, our database-inspired method leverages the power of R* trees for data aggregation and granulation. This approach makes it possible for the model to be implemented as a part of a Database Management System (DBMS), creating an opportunity to generate a scalable model as compared to the state-of-the-art.


The key factors that motivate this research include:

\noindent \textbf{Query response time:} A query requesting an estimation of the target variable at the current time should be responded to within a short time, i.e. it should have low latency.

\noindent \textbf{Accuracy of predictions:} The fast response needs to produce accurate predictions in order to be useful for subsequent decision-making tasks.

\noindent \textbf{Training time:} Along with the accurate fast response, it is also important for a model to train faster in order to handle continuous data streams for time-sensitive systems.

\noindent \textbf{Size of trained model:} To be able to handle unbounded data streams and maintain computational efficiency, it is necessary that the model size does not continuously increase in size with each instance of incoming data.

    The remainder of the paper is organized as follows: Related works are discussed in Section \ref{sec:related work}. Section \ref{granulation} includes details about how the model handles static datasets, which we extend further in Section \ref{Sec:Iterative Granulation} to account for data streams. Section \ref{experiments} discusses the experimental setup, datasets, and performance indicators for how the proposed approach compares against other methods. Section \ref{conclusion} wraps up the paper with some concluding remarks and a discussion of possible future directions.

\section{Related work}
\label{sec:related work}

In the field of data stream mining, numerous new techniques are introduced each year. However, as highlighted in \cite{gomes2018}, there is a much larger emphasis on developing methods for classification compared to those designed for regression. In this section, we review some of the relevant works that focus specifically on the problem of data stream regression.
    

    Adaptive Model Rules (AMRules) \cite{almeida2013} was proposed as an interpretable rule-based stream learning algorithm. For any given dataset, AMRules maintains a set of rules that are used to provide predictions. With an incremental learning approach, AMRules dynamically updates the rule set based on necessity when new data arrives with the help of a Page-Hinkley test \cite{page1954}.
    Similarly to AMRules, our proposed model is also interpretable and may be considered as a form of rule-based model. 
    Online Regression Trees with Options (ORTO) \cite{ikonomovska2011speeding} is another data stream regression approach that uses Hoeffding-based trees  \cite{domingos2000} but introduces optional nodes to the tree when the splitting decisions are ambiguous.
    Fast Incremental Model Trees with Drift Detection (FIMT-DD) \cite{ikonomovska2011learning} attempts to construct and update decision trees from data streams. While fast and effective, FIMT-DD can only process single instances, and is unable to handle batch data which may be necessary for handling large-scale, diverse data streams.
    Another related work is Adaptive Random Forests for Regression (ARF-Reg) \cite{gomes2018} that constructs a random forest with a modified version of FIMT-DD as base learners. This ensemble learning approach makes it more accurate but at the cost of evaluation time.
    In more recent works, Self-Optimizing K-Nearest Leaves (SOKNL) \cite{sun2022} extends the Adaptive Random Forest algorithm. In the prediction phase, instead of utilizing all the trees in the forest, SOKNL proposes to use only \textit{k} trees that cover instances closest to the queried instance. Though this can improve the accuracy, the evaluation time is even more than ARF, which creates problems for processing fast data streams.
    
    

    A key limitation of these works is that they all process a stream in an instance-by-instance manner. For large-scale applications, where there are multiple sources of a data stream, resulting in the arrival of millions of instances at the server every second, it would be beneficial to have a model that can process all instances that are available at once, i.e. in a \textit{batch learning} approach. Although the methods mentioned earlier are fairly fast, they can still fall short with a large enough influx of data due to the fact that they process data one instance at a time. 
    As mentioned in \cite{ikonomovska2011learning}, existing batch learning approaches require storing all the instances in the main memory for processing. This is challenging when dealing with data streams due to the high volume of continuous influx of data. Due to the large and ever-increasing volume of data, batch learning processes also take a long time to train. Several attempts have been made to speed up learning on large data such as \cite{dobra2002} and \cite{Vogel2007}. However, the memory requirements pose a major challenge in the context of deploying them for data stream learning. Our proposed method processes the incoming data in batches but does not require all the historical instances to be maintained in the main memory, as it identifies only the recent and relevant data and discarding all other instances. As we will demonstrate, this makes our model faster than the state-of-the-art online models by an order of magnitude. Moreover, since our method relies on a batch learning approach, it can also be utilized in industries that have multiple data stream sources.


\section{Data Granulation for Regression} \label{granulation}


In this section, we present a specific aggregation or \textit{data granulation} approach which systematically balances aggregation and fidelity. In Section \ref{Sec:Iterative Granulation} we leverage its inherent properties to create a novel stream regression method using iterative data stream granulation.


We begin with a dataset $\mathcal{D} = \{(\mathbf{x}_i, \mathbf{y}_i)\}$  $(i = 1, 2, ..., n)$, where $n$ may potentially be unbounded as in the case of a data stream. The dataset is comprised of input vectors $\mathbf{x} \in \mathcal{X}$, where $\mathcal{X}$ is a $d_x$-dimensional feature space, and the target variable $\mathbf{y} \in \mathcal{Y}$, where $\mathcal{Y}$ is a $d_y$-dimensional target variable space. The goal of our stream regression problem is to predict $\mathbf{y}$ given the input vector $\mathbf{x}$. In the proposed approach for stream regression, the first step in predicting the target variable $\mathbf{y}$ given the feature vector $\mathbf{x}$  is to determine a set of granules $\mathcal{G}$ that are (a) representative of the underlying dataset $\mathcal{D}$, and (b) less numerous than the data themselves. 
Conceptually, a granule refers to a hyper-rectangle in feature space that contains a specific set of proximate data points. The property of these proximate data points is that the mean value $\bar{\mathbf{y}}$ of their target variables (an aggregate characteristic) is most representative of the values of the set of target variable $\mathbf{y}_i$ associated with each of the individual data points. More specifically, the granule creation process is guided by the concept of Allan variance (AVAR) or 2-sample variance \cite{allan1966}.
The use of AVAR in granule formation helps maintain a balance between variance and bias in the target variable. For further details on the application of AVAR, the reader may refer to the works of \cite{haeri2021, haeri2022optimal, sinanaj2022, Jerath2018, allan1966}. More specifically, for additional details pertaining to the adaptive granulation method, the reader is directed towards \cite{haeri2023}.

A granule $g_j \in \mathcal{G}$ is defined as a tuple $(\mathcal{D}_j, \textbf{x}_j^\text{min}, \textbf{x}_j^\text{max}, \bar{\mathbf{y}}_j)$, where   $\mathcal{D}_j = \{(\mathbf{x}_i, \mathbf{y}_i)_j\}$ such that $\mathcal{D}_j \subset \mathcal{D}$, i.e. it is the subset of $\mathcal{D}$ which contains the instances covered by the $j^{th}$ granule, $\textbf{x}_j^{\text{min}}$ and $\textbf{x}_j^{\text{max}}$ are $d_x$-dimensional vectors representing the minimum and maximum bounds of the granule or hyper-rectangle in the $d_x$-dimensional feature space, and  $\bar{\textbf{y}}_j$ is a $d_y$-dimensional vector representing the average of the target variable vectors of the elements of $\mathcal{D}_j$. 


The granules can be used as the regression model.  Specifically, we can use the $\bar{\textbf{y}}_j$ value of a granule $g_j$ as the prediction for any query that falls within the region covered by $g_j$ in the $d_x$-dimensional hyperspace. A $d_x$-dimensional query point $\mathbf{x}_q \in \mathcal{X}$ is considered to be covered by a granule $g_j$ if and only if 
$\mathbf{x}_q \geq {\mathbf{x}}^\text{min}_j \:\: \mbox{and}  \:\:\mathbf{x}_q \leq {\mathbf{x}}^\text{max}_j$
and given this query point, the granulation-based regression model provides the predicted target variable as $\hat{\mathbf{y}}_q = \mathbf{\bar{y}}_j$. Algorithm \ref{al:granule model} shows how a list of granules can be used to respond to a query $\mathbf{x}_q$. Notice on lines 19-21 that if more than one granule covers a query point, then a mean of their respective $\bar{y}$ values will be used as the prediction, i.e.,
\begin{align}
\hat{\mathbf{y}}_q = \dfrac{1}{|\mathcal{G}_C|}\sum_{g_j\in \mathcal{G}_c}\bar{\mathbf{y}}_j
\label{eq: multiple granules}
\end{align}
If no such covering granules are found, then the granule that is closest to the query point is considered the covering granule (lines 16-18). The distance $d(\mathbf{x}_q, g_j)$ between a point $\mathbf{x}_q$ and a granule $g_j$ is defined as the Euclidean distance between $\mathbf{x}_q$ and the vertex of the hyperrectangle (granule) that is closest to the point, as follows: 
\begin{equation}
d(\mathbf{x}_q, g_j) = \sum_{d_x}\text{min}(\mathbf{x}_q - {\mathbf{x}}^\text{max}_j,\:\: \mathbf{x}_q - {\mathbf{x}}^\text{min}_j)^2
\label{eq: closest granule}
\end{equation}
Since the granulation process utilizes the nodes of an R* tree as granules\cite{haeri2023}, there exists a possibility that some of the granules will overlap in the feature space. While this overlap may affect the accuracy of predictions, this is offset by the fast response time advantage offered by R* trees, which facilitate the deployment of our method on database systems.

\begin{algorithm}[t]
    \small
    \begin{algorithmic}[1]
    \REQUIRE Set of granules ($\mathcal{G}$), query point ($\mathbf{x}_q$), number of dimensions ($d_x$)
    
    \ENSURE Predicted target variable ($\hat{\mathbf{y}}_q$)

    \STATE \texttt{CG} $\gets \phi$ \hspace{1.2em}$\triangleright$ create empty set \texttt{CG} of covering granules
    \STATE $\texttt{sum} \gets 0$
    \FOR{$\texttt{g} \in \mathcal{G}$} 
        \STATE \texttt{flag} $\gets$ \texttt{True}
        \STATE $\texttt{i} \gets 0$
        \WHILE{$\texttt{i} \leq d_x$}
            \IF{$\texttt{x\textsubscript{q}(i)} > \texttt{max(g.x(i))}$ \OR $\texttt{x\textsubscript{q}(i)} < \texttt{min(g.x(i))}$}
                \STATE \texttt{flag} $\gets$ \texttt{False}
            \ENDIF
                \STATE \texttt{i} $\gets$ \texttt{i} + 1
        \ENDWHILE
        \IF{$\texttt{flag} == \texttt{True}$}
                \STATE $\texttt{CG.insert(g)}$
        \ENDIF
    \ENDFOR

    \IF{$\texttt{CG} == \phi$}
        \STATE \texttt{CG.insert(FindClosestGranule($\mathcal{G}$, $x_q$))}
    \ENDIF
    
    \FOR{$\texttt{g} \in \texttt{CG}$}
        \STATE \texttt{sum} $\gets \texttt{sum} + \texttt{g}.\overline{\texttt{y}}$ 
    \ENDFOR

    \STATE \texttt{\^{y}\textsubscript{q}} = $\texttt{sum} / \texttt{|CG|}$
    \RETURN \texttt{\^{y}\textsubscript{q}}
    \end{algorithmic}
    
    \caption{Querying the granule-based model}
    \label{al:granule model}
\end{algorithm}

    

    

    


\section{Iterative Forgetting as a Datastream Regression Model}
\label{Sec:Iterative Granulation}

In this section, we discuss the methodology for extracting recent and relevant granules and using them to predict the target variable. Before using the proposed iterative granulation method we perform some standard pre-processing steps such as (a) outlier rejection, i.e. eliminating data points whose values for individual features are $\mu \pm 3\sigma$. For data streams, the value of $\mu$ and $\sigma$ can be calculated by maintaining a running sum and sum of squares, and (b) creating a single feature to capture temporal aspects of the data, such as by reformatting the temporal feature variables (e.g., separate features for month, day, hour, etc.) into a single feature to create a dataset representative of data streams. We now focus on (a) how \textit{relevant} information can be extracted in the form of recent granules, making it feasible to discard outdated information, and (b) how we can iteratively perform recent granule extraction over batches of incoming data, i.e. for data streams.


\subsection{Recent granule extraction}\label{rg extraction}
While granules partition the proximate data points optimally using Allan variance \cite{haeri2023}, not all of them may be required to respond to a query point that requests the value of the target variable at the current time. In this subsection, we show how we label the granules that contain the newest information in the entire feature space as \textit{Recent Granules}, and all other granules are labeled as \textit{Outdated Granules}.


When a set of granules is projected onto the current time plane, a granule that is entirely overlapped by other granules with a higher value of $\textbf{x}_j^\text{max}$ for the temporal feature is considered to contain past or `old' information that is not relevant for prediction purposes. Conversely, a granule that is not overlapped by any other granules with a higher value of $\textbf{x}_j^\text{max}$ of the temporal feature is considered to contain newer information. In simple terms, recent granules are the ones that are visible from the hyperplane located at current time $T \geq \mbox{max}(x^t_i)$, where $x^t$ represents the temporal feature of the data $\mathcal{D}$ and $T$ is the temporal feature axis. Fig. \ref{fig:rg_vs_nrg} provides an example of this in a 2-dimensional space. The yellow granules are visible when observed from the current time plane and contain newer information for their respective regions in the feature space, making them recent granules. The gray granules are obstructed by other granules and are not visible, making them outdated granules that contain older information for their respective regions in the feature space.

We claim that for any queries requesting the estimate of the target variable $\mathbf{\hat{y}}_q$ at current time $T \geq \text{max}(x_t))$, the recent granules are best suited for providing these predictions and the outdated granules only contain either redundant or expired information. This is a reasonable claim since (a) each granule is created using the optimal aggregation scale of the data, which implies that the optimal scale of relevance is captured for the temporal attribute as well \cite{haeri2021, haeri2022optimal}, and (b) at current time $T$, the recent granules effectively cover the $d_x -1$ dimensional feature space of the data.
 With this reasoning, we can remove the outdated granules and the instances that reside within them, and keep only the recent granules and the data that reside within them, which will be referred to as \textit{recent data} in the remaining.

Algorithm \ref{al:rg extraction} shows an efficient way of extracting the set of recent granules $\mathcal{G}_R$. For each instance of the input data, finding the granules that cover the instance in $d_x -1$ dimensional space (disregarding the temporal feature) and picking the granule with the highest $\textbf{x}_j^\text{max}$ value for the temporal dimension among these covering granules results in the same set of recent granules as depicted in the Fig. \ref{fig:rg_vs_nrg}. The covering granules in $d_x-1$ dimensional space can be extracted by returning the set $\mathcal{G}_C$ instead of the predicted value in Algorithm \ref{al:granule model}. The set of recent granules ($\mathcal{G}_R$) extracted with the help of Algorithm \ref{al:rg extraction} will act as the new model for regression.

\begin{algorithm}
\small
    \begin{algorithmic}[1]
    \REQUIRE Set of granules ($\mathcal{G}$), number of dimensions ($d_x$), index of the temporal dimension ($x_t$)
    \ENSURE Set of recent granules ($\mathcal{RG}$)

    \STATE $\texttt{RG} \gets \phi$ \hspace{2em}$\triangleright$ create empty set \texttt{RG} of recent covering granules
    
    \FOR{$\texttt{g} \in \mathcal{G}$}
        \FOR{$\texttt{e} \in \mathcal{D}_g$}
            \STATE $\mathcal{CG} \gets$ \texttt{CoveringGranules($\mathcal{G}$, e, $d_x$, $x_t$)}
            \STATE sort $\mathcal{CG}$ in the decreasing order of $\text{max}[x_t]$
            \STATE \texttt{CRG} $\gets$ $\mathcal{CG}[0]$
            \STATE \texttt{RG.insert(\texttt{CRG})}
            \IF{\texttt{CRG} == \texttt{g}}
                \STATE \textbf{break}            
            \ENDIF
        \ENDFOR        
    \ENDFOR
        
    \RETURN \texttt{RG}
    \end{algorithmic}    
    \caption{Recent granules extraction}
    \label{al:rg extraction}
\end{algorithm}

\subsection{Iterative Data Stream Granulation}\label{iterative forgetting}
The recent granule extraction explained in the previous section utilizes a granulation method that assumes that all the data is available beforehand, which is not true in the context of data streams. This section explains how recent granule extraction can be applied to data streams by continuously processing the data that \textit{is} available.

For recent granules extraction in continuous data streams, we consider the data to be accumulated in a buffer and processed in batches. A batch could be defined in terms of a specific number of instances or in terms of the amount of time elapsed since the last batch was processed. Both definitions have their own benefits: When the frequency of receiving data is very high, it would be better to choose the data accumulated in a fixed amount of time as a batch, and when the frequency of receiving data is low or the data is received intermittently, it would be better to choose the batch size based on the number of instances in the buffer.

\begin{figure*}
            \centering
            \includegraphics[width = \linewidth]{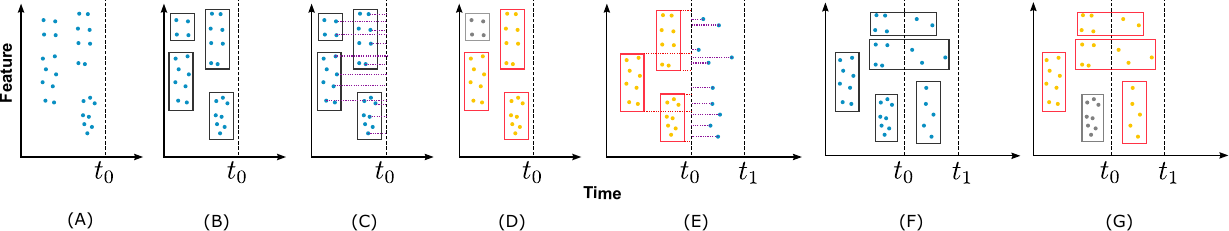}
            \caption{Iterative Forgetting process. Details for each of the steps are provided in Section \ref{Sec:Iterative Granulation}.B.}
            \label{fig:if-process}
    \end{figure*}

    The iterative forgetting process is explained below and visualized in Fig. \ref{fig:if-process}:

    \noindent \textbf{Initialization:} We initialize the set of recent data $\mathcal{D}_R$ and the set of recent granules $\mathcal{G}_R$ as empty sets. We will maintain and update these sets at each iteration of the iterative forgetting process.

    \noindent \textbf{Step A}: The first batch $b_1$ of data is obtained from the incoming datastream, i.e. the data has just started streaming.

    \noindent \textbf{Step B}: The first batch of data is processed and the granules are created using proximate data points and the Allan variance-based adaptive granulation approach \cite{haeri2023}.
     
    \noindent \textbf{Step C}: The instances within the batch are used as queries to find the covering granules. This means that we create projections of the feature data points $(\mathbf{x}_i)_j$ onto the current time plane. In Fig. \ref{fig:if-process}(c), the projections are shown for only a few data points.

    \noindent \textbf{Step D}: We identify which of the projections onto the current time plane (obtained in Step C) are unobstructed by any other granules, i.e. which data points are `visible' from the current time plane. The recent granules $\mathcal{G}_R$ (red) and recent data $\mathcal{D}_R$ (yellow) are extracted by choosing the most recent (`visible') granule from the covering granules for each query point. The outdated granules and outdated data (gray) are removed from consideration. 

    \noindent \textbf{Step E}: The set of recent granules $\mathcal{G}_R$ is used as a datastream regression model to respond with predictions $\mathbf{\hat{y}}_q$ to the queries made by data streaming in during the creation of the next batch. The prediction is provided by finding the recent granules $\mathcal{G}_R$ that cover the query points and using these as the input set of granules $\mathcal{G}$ in Algorithm 1.

    \noindent \textbf{Step F}: The true measurements $\mathbf{y}_i$ for the feature data points that were queries in Step E are now available as constituents of the next batch $b_2$ ($t_0 < b_2 \le t_1$). The recent data $D_R$ and the data in the new batch $b_2$ are combined and used to create new granules similar to Step B.

    \noindent \textbf{Step G}: A \textit{new} set of recent granules $\mathcal{G}^\text{new}_R$ and recent data $\mathcal{D}^\text{new}_R$ are extracted by repeating Steps C and D. 

        \begin{figure}
            \centering
            \includegraphics[width = 0.8\linewidth]{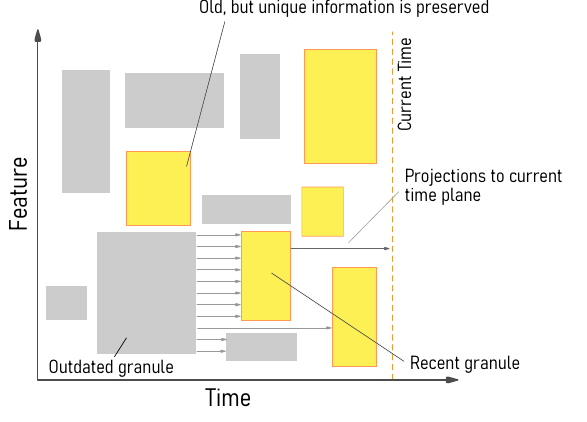}
            \caption{Recent granules (yellow) and outdated granules (gray)}
            \label{fig:rg_vs_nrg}
\end{figure}

    \vspace{0.5em}
    Steps C-G are performed iteratively as part of the recent granule extraction algorithm. Notice that after the first batch, all the other batches are processed after merging them with the instances that are stored in the recent data at that time, and in every iteration, a new set of recent data and recent granules are extracted. This ensures that every time a new batch is processed, the data that was found to be relevant and recent during the previous batch processing will be checked again. If any instances from the previously extracted recent granules are no longer relevant, then those will not be included in the new set of recent data. Concurrently, if there are some instances that were found to be relevant in a previous update and are still relevant, the recent granules extraction algorithm will make sure that those are a part of the new set of recent data. This scenario is visualized in Fig. \ref{fig:rg_vs_nrg} and can also be seen in Fig. \ref{fig:if-process} (G). At any given time, the set of recent granules ($\mathcal{G}_R$) is used as a model to address the queries, following the Algorithm \ref{al:granule model} (see Fig. \ref{fig:if-process} (E)). By continuously processing data as it becomes available, the iterative forgetting process that is proposed here ensures the extraction of relevant and recent granules for addressing queries accurately in real-time. We can now examine the efficacy of the proposed iterative datastream granulation approach.
    

\section{Experiments and Results} \label{experiments}
In this section, we compare the performance of our model against the four state-of-the-art stream regression models (FIMT-DD \cite{ikonomovska2011learning}, AMRules \cite{almeida2013}, ORTO \cite{ikonomovska2011speeding}, and ARF \cite{gomes2018}) for three synthetic and six real-world datasets. Our performance criteria, as mentioned in the introduction, include (a) evaluation time, which is the amount of time it takes to train and test each model for each dataset, (b) prediction accuracy, as measured by the Mean Absolute Error (MAE) and Root Mean Squared Error (RMSE), and (c) model size, determined as the amount of memory required to store the model throughout the training phase and measured in bytes. 
A larger model size could lead to increased transmission latency between systems, potentially causing time-sensitive systems to operate on outdated information, thereby affecting their performance. All the experiments are performed on a Windows 10 system, with 40GB of RAM (the Java execution was limited to only 10GB through the IDE) and a Ryzen 7 5800HS processor with 8 cores.

We utilize the \textit{test-then-train} approach, where each new instance is first used to test the model and then to train the model. For comparison with the state-of-the-art models, we use the implementations present in Massive Online Analysis (MOA) library \cite{bifet2010moa}. Similarly to the MOA library, our model is implemented in Java 8 (OpenJDK 1.8).
We have taken the following measures to make sure that the comparison is fair:

\noindent\textbf{Ensuring fairness in evaluation time:} In MOA, `sampling frequency' denotes the interval of instances for reporting model performance metrics. For instance, a frequency of 100 means performance is reported after processing every 100 instances. Notably, frequent reporting significantly increases evaluation time. To ensure fair comparison, we set the sampling frequency higher than our datasets' sizes, ensuring performance is reported only after processing the entire dataset.

\noindent\textbf{Ensuring fairness in model size comparison:} For memory comparisons, experiments vary the sampling frequency from 100 to 10000, depending on dataset size. We consider only the model size values from these results, as the evaluation time is inflated due to the sampling frequency.

\noindent\textbf{Averaged results:} The reported results for the evaluation time are averaged over five runs for each case.

\noindent\textbf{Outlier rejection:} The outliers detected by our model are still included in measuring the MAE and RMSE and are only excluded from the granulation and recent granule extraction phase since the other models are also including the estimates for each instance in the provided dataset.

\subsection{Synthetic data}
\begin{figure}
        \centering
        \includegraphics[width = 0.8\linewidth]{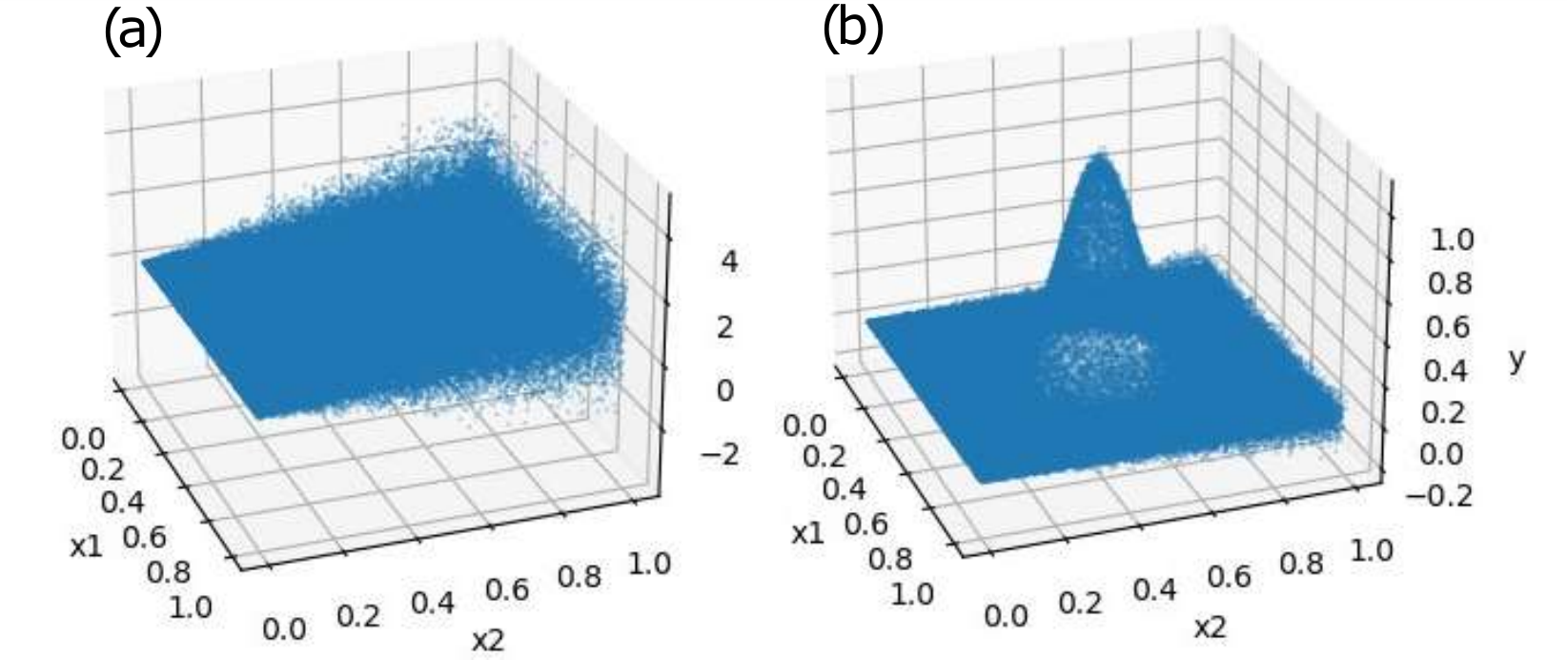}
        \caption{Synthetic data (a) noise varying, and (b) noise and parameter varying.}
        \label{fig:noise-data}
    \end{figure}

    For synthetic scenarios, we have taken two datasets from \cite{haeri2023} that represent changes in underlying pattern complexity and changes in noise over attribute space. The dataset with pattern change is also corrupted with varying noise, similar to the noise-varying dataset, and the attribute along which noise varies is considered the temporal attribute for both datasets. Both of these datasets contain 100,000 instances. Fig. \ref{fig:noise-data}(a) and (b)
    visualize these datasets.

\begin{figure}
            \centering
            \includegraphics[width = 0.8\linewidth]{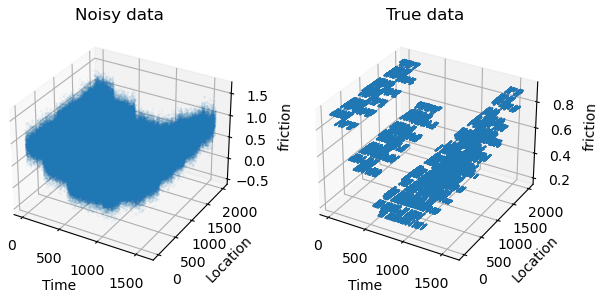}
            \caption{Simulated road friction data (SRFD). Noisy input (left), True underlying values (only for our understanding) (right)}
            \label{fig:srfd}
    \end{figure}

    Moreover, we have considered one additional dataset, \textit{Simulated Road Friction Data (SRFD)} that simulates a scenario with multiple vehicles transmitting their locations, road friction measurement at that location (noisy measurement), and a timestamp \cite{GAO2022}. Fig. \ref{fig:srfd} visualizes this dataset where \texttt{sim\_wall\_time} refers to the timestamp, \texttt{cg\_station} refers to the location (only used for visualization. The models are trained with two attributes for the location that represent the East and North coordinates of the vehicle). The left figure shows the noisy data that is used as input to the models and the right figure shows the true underlying values of each instance that is just used to show how the measurements change with time and with location. This dataset consists of 647,400 records.

\subsection{Real-world data}
    For real-world scenarios, we have utilized seven datasets that are as follows: (1) Abalone (4100 instances) \cite{abalone1995}; (2) Air Quality (9300 instances) \cite{air_quality2016}; (3) Bike (10k instances) \cite{fanaee2014}; (4) Metro Interstate Traffic Volume (48k instances) \cite{metro_traffic2019}; (5) NYC taxi trip duration (1.4M instances) \cite{nyc_taxi};  (6) Wind speed (5600 instances) \cite{Wind2022}.


\subsection{Results}

    \begin{figure}
            \centering
            \includegraphics[width=\linewidth]{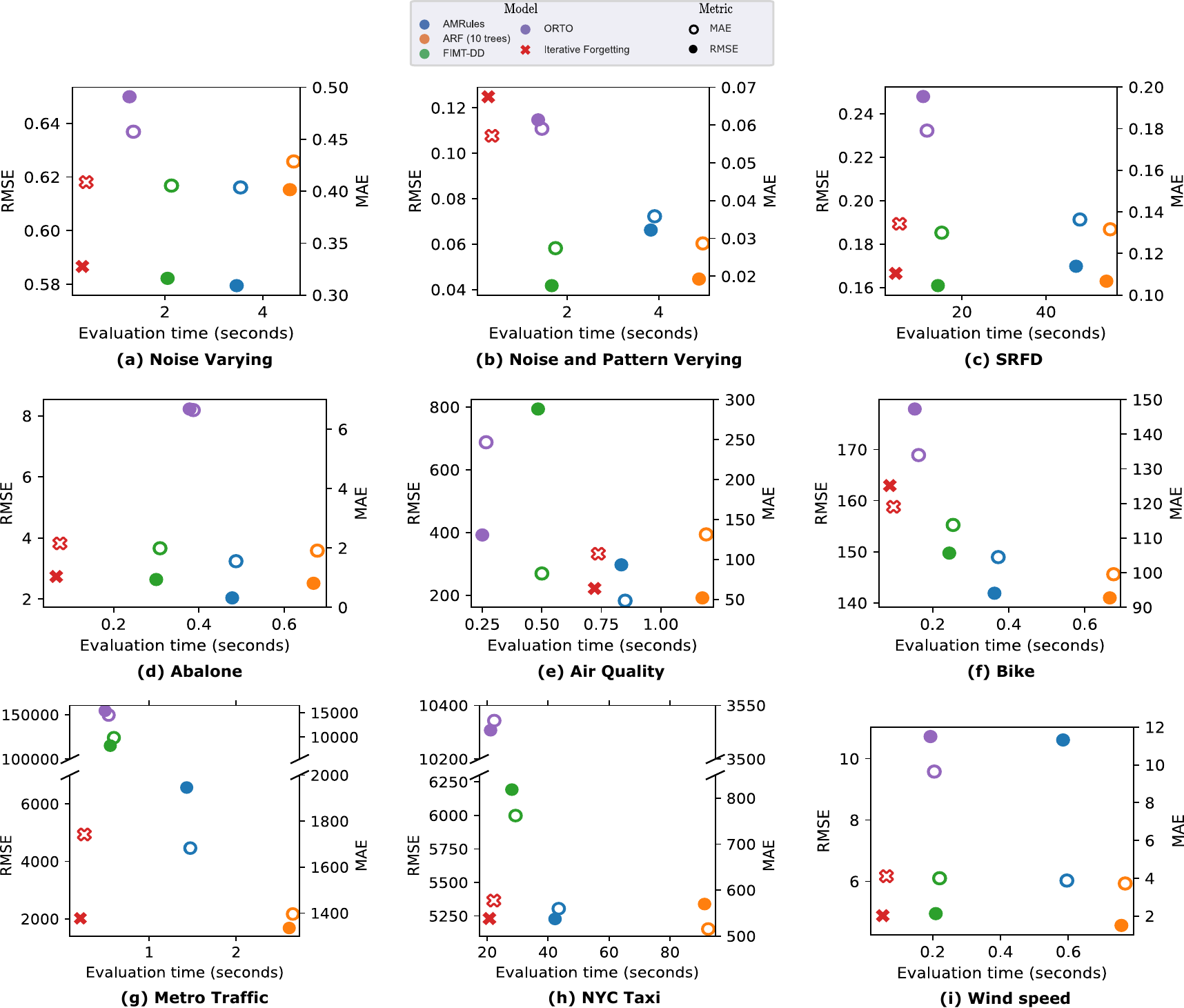}
            \caption{Evaluation time (seconds) and accuracy (MAE, RMSE) comparison}
            \label{fig:res-time}
    \end{figure}

    \begin{figure}
            \centering
            \includegraphics[width=\linewidth]{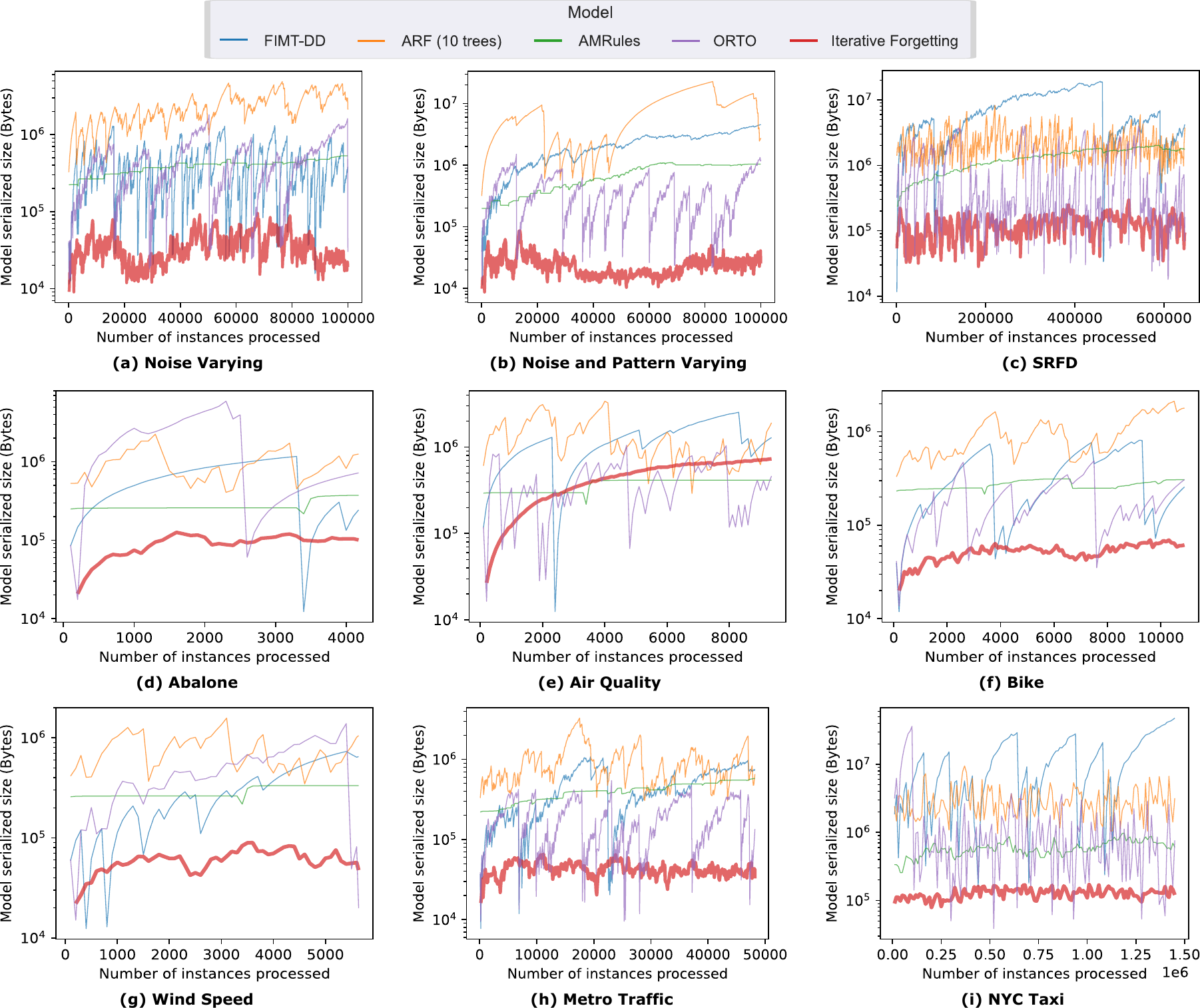}
            \caption{Model size (bytes) comparison}
            \label{fig:res-size}
    \end{figure}

    Fig. \ref{fig:res-time} shows the performance comparison for the datasets. The x-axis represents the evaluation time of the model in seconds and the y-axis shows the accuracy of the model in terms of  MAE and RMSE. The results show that other than the Air Quality dataset, Iterative Forgetting is consistently faster than other methods while maintaining comparable accuracy for most datasets. It can be seen that among the state-of-the-art methods, ORTO is generally the fastest and our method can be up to 6 times faster than ORTO (e.g., Fig. \ref{fig:res-time}(d)) while maintaining significantly less error than ORTO. The most accurate method among state-of-the-art is ARF, which performs well in both MAE and RMSE for most datasets. Our method can be more than 10 times faster than ARF (refer to the results of the Wind speed and the Metro traffic datasets.) which makes sense as ARF, being an ensemble learning method, trades evaluation time for better accuracy.

    Fig. \ref{fig:res-size} shows how the amount of space a particular model would require to be stored varies throughout the evaluation. The results indicate that the Iterative Forgetting model consistently requires less size than the other models except for the air quality dataset. In the Air Quality dataset, the model size chart suggests that no instances are being forgotten, which would explain why Iterative Forgetting is slower to process in this dataset. The evaluation time in Iterative Forgetting is strongly related to the size of the model during the training phase. If a model is large, it will take more time to extract granules from the model and if no information is considered outdated, then the model size continuously keeps increasing, resulting in higher update times for the consecutive batches.

    In general, Iterative Forgetting can outperform the state-of-the-art methods in terms of the evaluation time and model size while introducing a marginal increase in the error.

\section{Concluding remarks and future work} \label{conclusion}

We have proposed an interpretable model that exhibits faster performance compared to existing methods while maintaining comparable accuracy. Our model is represented as a set of rectangles, where each rectangle is associated with a target attribute value. This inherent structure makes our model highly interpretable, enabling users to easily understand the reasoning behind predictions.

Additionally, since our model is developed with R* trees as a foundation, it can be implemented on the database level. Its lower memory requirements also allow for efficient storage of the model in a database after each update, enabling queries to be addressed directly by the database system. The proposed method can be seamlessly integrated into existing database systems, offering a practical and scalable solution for real-time query processing and decision-making tasks. Further exploration of utilizing our model's database integration capability for distributed systems is discussed in \cite{pakala2023}.
Overall, our work is expected to provide significant benefits in various domains where interpretable models, fast processing, memory efficiency, and distributed systems are crucial considerations.

\balance

\bibliographystyle{IEEEtran}
{\small
\bibliography{references}}

\end{document}